\documentclass{sig-arxiv}
\usepackage{lipsum}
\usepackage{booktabs}
\usepackage{multirow}
\usepackage{color}
\usepackage{pgfplots}
\usepackage{microtype}
\usepackage{tikz}
\usepackage{array}
\usepackage[tableposition=top]{caption}
\usepackage{float}

\usetikzlibrary{fit,shapes.geometric}
\newcommand*\samethanks[1][\value{footnote}]{\footnotemark[#1]}
\newcounter{nodemarkers}

\newcolumntype{L}[1]{>{\raggedright\let\newline\\\arraybackslash\hspace{0pt}}m{#1}}
\newcolumntype{C}[1]{>{\centering\let\newline\\\arraybackslash\hspace{0pt}}m{#1}}
\newcolumntype{R}[1]{>{\raggedleft\let\newline\\\arraybackslash\hspace{0pt}}m{#1}}

\usepackage{graphicx}
\definecolor{bblue}{HTML}{4F81BD}
\definecolor{rred}{HTML}{C0504D}
\definecolor{ggreen}{HTML}{9BBB59}
\definecolor{ppurple}{HTML}{9F4C7C}
\usepackage{adjustbox}
\begin{document}

% \CopyrightYear{2016} 
% \setcopyright{acmcopyright}
% \conferenceinfo{ICMR'16,}{June 06-09, 2016, New York, NY, USA}
% \isbn{978-1-4503-4359-6/16/06}\acmPrice{\$15.00}
% \doi{http://dx.doi.org/10.1145/2911996.2912061}

%Authors, replace the red X's with your assigned DOI string.

\clubpenalty=10000 
\widowpenalty = 10000

%\title{Local CNN Features using Bag of Words for Instance Search}
\title{Bags of Local Convolutional Features for Scalable Instance Search}


% Xavi - Double blind review submission

\numberofauthors{2}
\author{
\alignauthor Eva Mohedano\thanks{denotes equal contribution.}, Kevin McGuinness and Noel E. O'Connor \\
\affaddr{Insight Center for Data Analytics} \\
\affaddr{Dublin City University} \\
\affaddr{Dublin, Ireland} \\
\email{eva.mohedano@insight-centre.org}
\alignauthor Amaia Salvador\samethanks, Ferran Marqu\'es, \\and  Xavier Gir\'o-i-Nieto\\
\affaddr{Image Processing Group} \\
\affaddr{Universitat Politecnica de Catalunya} \\
\affaddr{Barcelona, Catalonia/Spain} \\
\email{\{amaia.salvador, xavier.giro\}@upc.edu}
} 

% \numberofauthors{6}
% \author{
% Eva Mohedano\thanks{denotes equal contribution.}
% \and Amaia Salvador\samethanks
% \and Kevin McGuinness
% \and Xavier Gir\'o-i-Nieto
% \and Noel O'Connor
% \and Ferran Marqu\'es
% }

\maketitle
\begin{abstract}
%Image representations extracted from convolutional neural networks (CNNs) have been shown to outperform hand-crafted features in multiple computer vision tasks, such as visual image retrieval. 

This work proposes a simple instance retrieval pipeline based on encoding the convolutional features of CNN using the bag of words aggregation scheme (BoW). Assigning each local array of activations in a convolutional layer to a visual word produces an \textit{assignment map}, a compact representation that relates regions of an image with a visual word. We use the assignment map for fast spatial reranking, obtaining object localizations that are used for query expansion. We demonstrate the suitability of the BoW representation based on local CNN features for instance retrieval, achieving competitive performance on the Oxford and Paris buildings benchmarks. We show that our proposed system for CNN feature aggregation with BoW outperforms state-of-the-art techniques using sum pooling at a subset of the challenging TRECVid INS benchmark.
\end{abstract}

%\category{H.3.1}{Information Storage and Retrieval}{Content Analysis and Indexing}
%A category including the fourth, optional field follows...
%\category{I.4.7}{Image Processing and Computer Vision}{Feature Measurement}[Feature representation]

% <concept_id>10002951.10003317.10003371.10003386.10003387</concept_id>
% <concept_desc>Information systems~Image search</concept_desc>
% <concept_id>10002951.10003317.10003365.10003366</concept_id>
% <concept_desc>Information systems~Search engine indexing</concept_desc>
% <concept_significance>100</concept_significance>

% \ccsdesc[500]{Information systems~Image search}
% \ccsdesc[100]{Information systems~Search engine indexing}

% Check categories here: http://www.acm.org/about/class/ccs98-html
\terms{Algorithms}

% Your general terms must be any of the following 16 designated terms: Algorithms, Management, Measurement, Documentation, Performance, Design, Economics, Reliability, Experimentation, Security, Human Factors, Standardization, Languages, Theory, Legal Aspects, Verification.


\section{Introduction}
\label{1_intro}

% Visual search
%Visual image retrieval is concerned with organizing and structuring datasets of images based on their visual content. 

%The proliferation of ubiquitous cameras in the last decade has motivated researchers in the field to push the limits of visual search systems with scalable yet effective solutions. 

% CNNs
Convolutional neural networks (CNNs) have been demonstrated to produce global representations that effectively capture the semantics in images. Such CNN-based representations ~\cite{babenko2015,neuralcodes,cnnofftheshelf,razavian2015,tolias2015,xie2015image} have improved upon the state-of-the-art for instance retrieval.

\begin{figure}
  \includegraphics[width=\columnwidth]{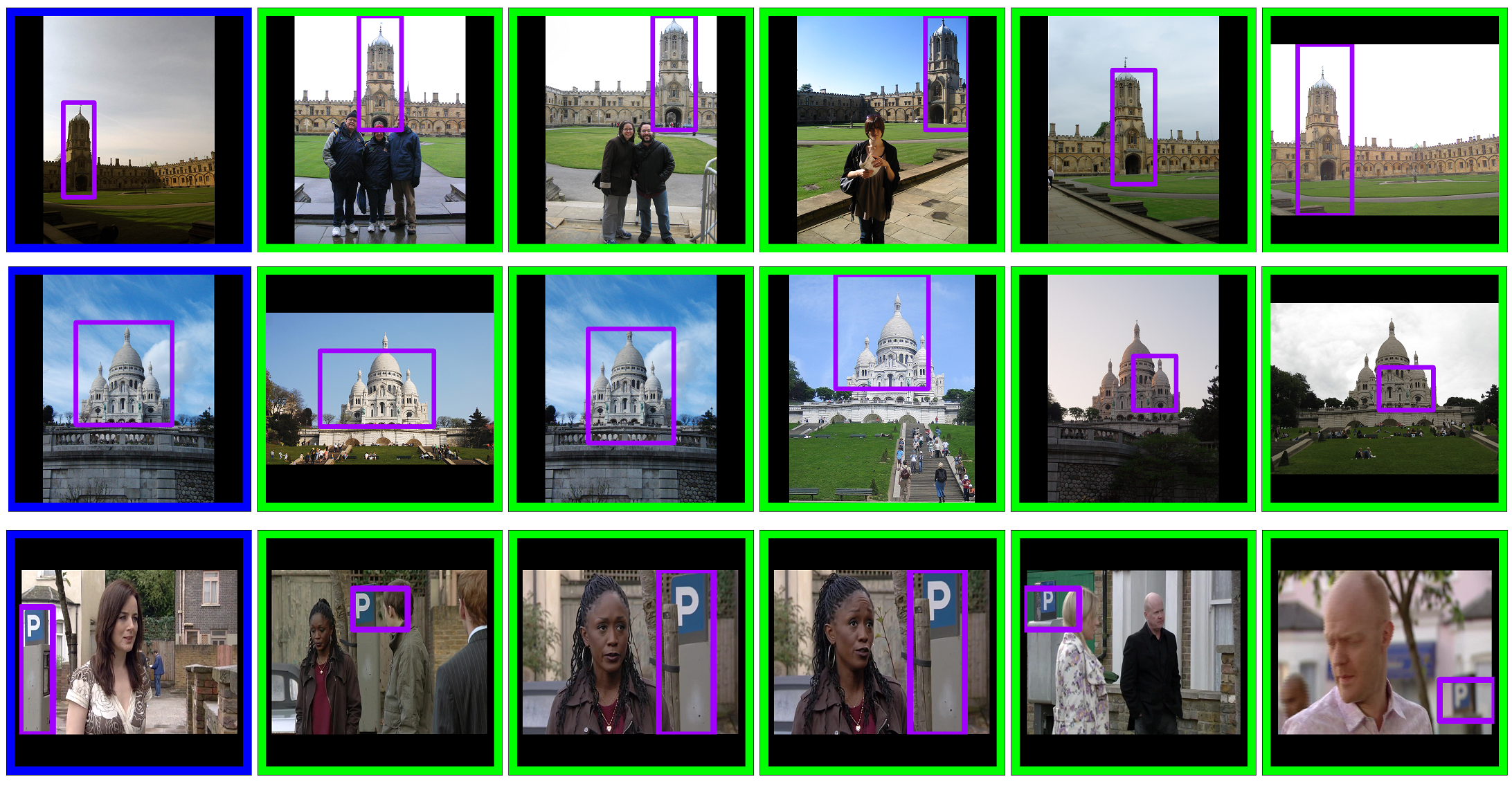}
  \caption{Examples of the top generated rankings for queries of three different datasets: Oxford (top), Paris (middle), TRECVid INS 2013 (bottom).}
  \label{fig:rankings}
\end{figure}

% Limitations for Instance search

Despite CNN-based descriptors performing remarkably well in retrieval benchmarks like Oxford and Paris Buildings, state-of-the-art solutions for more challenging datasets such as TRECVid Instance Search (INS) have not yet adopted pipelines that depend solely on CNN features. Current INS systems~\cite{nguyen2015query,zhang2015topological,zhou2014practical, zhu2012large} are still based on aggregating local hand-crafted features using bag of words (BoW) encoding~\cite{bow} to produce very high-dimensional sparse image representations. Such high-dimensional sparse representations are more likely to be linearly separable, while having relatively few non-zero elements makes them efficient both in terms of storage and computation. 

%Sparse representations can handle varying information content, and are less likely to interfere with one another when pooled. From an information retrieval perspective, sparse representations can be stored in inverted indices, which facilitates efficient selection of images that share features with a query. Furthermore, there is considerable evidence that biological systems make extensive use of sparse representations for sensory information~\cite{lennie2003cost, vinje2000sparse}. Empirically, sparse representations have repeatedly demonstrated to be effective in a wide-range of vision and machine learning tasks.

% Despite the fact that CNN-based descriptors perform remarkably well for Content Based Image Retrieval (CBIR) (i.e. when the task is to find similar images or images with the same content as the given query), they still have not been adopted to solve other image retrieval tasks, such as instance search, where the goal is to find images containing a specific instance (e.g. an specific object, person, etc.). 

%
% NOTE FROM KEVIN:
% Is the above sentence strictly true? We ourselves have 
% used deep CNNs for instance search on TRECVid in the 
% past! Let's also review the TRECVID workshop papers 
% for this year to see what people did.
%

% XAVI:
Many successful image retrieval engines combine an initial highly-scalable ranking mechanism on the full image database with a more computational demanding yet higher-precision reranking mechanism (e.g. geometric verification) applied to the top retrieved items.

%This reranking mechanism often takes the form of geometric verification and spatial analysis~\cite{jegou2010improving, zhang2011image, mei2014multimedia,zhang2015topological}, after which the best matching results can be used for query expansion (pseudo-relevance feedback)~\cite{arandjelovic2012three,chum2007total}.

% Our work

Inspired by advances in CNN-based descriptors for image retrieval, yet still focusing on instance search, we revisit the BoW encoding scheme using local features from convolutional layers of a CNN. 
This work presents three contributions:

\begin{itemize}
\setlength{\itemsep}{0pt}
\item We propose a sparse visual descriptor based on a Bag of Local Convolutional Features (BLCF), which allows fast image retrieval by means of an inverted index. 

\item We introduce the assignment map as a new compact representation of the image, which maps pixels in the image to their corresponding visual words, allowing the fast composition of a BoW descriptor for any region of the image. 

\item We take advantage of the scalability properties of the assignment map to perform a local analysis of multiple image regions for reranking, followed by a query expansion stage using the obtained object localizations. 

\end{itemize}
Using this approach, we present a retrieval system that achieves state-of-the-art performance in several instance retrieval benchmarks.
Figure~\ref{fig:rankings} illustrates some of the rankings produced by our system on several datasets.

% Paper structure
%The remainder of the paper is structured as follows. Section~\ref{2_soa} presents related work. Section~\ref{3_bow} introduces the proposed framework for BoW encoding of CNN local features. Section~\ref{5_retrieval} explains the details of our retrieval system, including the local reranking and query expansion stages. Section~\ref{5_experiments} presents experimental results on three different retrieval benchmarks (Oxford Buildings, Paris Buildings, and a subset of TRECVid INS 2013), as well as a comparison to five other state-of-the-art approaches. Section~\ref{6_conclusions} summarizes the most significant results and outlines future work.
%
% KEVIN: if we are stuck for space, we can leave out the above paragraph
%

\section{Related Work}
\label{2_soa}

Many works in the literature have proposed CNN-based representations for image retrieval. Early works \cite{neuralcodes,cnnofftheshelf} focused on replacing traditional hand-crafted descriptors with features from fully connected layers of a CNN pre-trained for image classification. A second generation of works reported significant gains in performance when switching from fully connected to either sum \cite{babenko2015,kalantidis2015} or max \cite{razavian2015,tolias2015} pooled convolutional features. Our work shares similarities with all the former in that we use convolutional features extracted from a pre-trained CNN. Unlike these approaches, however, we propose using a sparse, high-dimensional encoding that better represents local image features.

Several works have tried to exploit local information in images by passing multiple image sub patches through a CNN to obtain local features from either fully connected~\cite{cnnofftheshelf, deepindex} or convolutional~\cite{gong2014multi} layers, which are in turn aggregated using techniques like average pooling~\cite{cnnofftheshelf}, BoW~\cite{deepindex}, or VLAD~\cite{gong2014multi}. Although many of these methods perform well in retrieval benchmarks, they require CNN feature extraction from many image patches, which slows down indexing and feature extraction at retrieval time. Instead, other works \cite{ng2015, netvlad} treat the activations of the different neuron arrays across all feature maps in a convolutional layer as local features. This way, a single forward pass of the entire image through the CNN is enough to obtain the activations of its local patches, which are then encoded using VLAD. Our approach is similar to the ones in~\cite{ng2015,netvlad} in that we also treat the features in a convolutional layer as local features. We, however, use BoW encoding instead of VLAD to take advantage of sparse representations for fast retrieval in large-scale databases.

%Following this approach, Ng et al.~\cite{ng2015} propose to use VLAD~\cite{vlad} encoding of features from convolutional layers to produce a single image descriptor. Arandjelovi et al.~\cite{netvlad} choose to adapt a CNN with a layer especially trained to learn the VLAD parameters. Our approach is similar to the ones in~\cite{ng2015,netvlad} in that we also treat the features in a convolutional layer as local features extracted at different locations in an image. We, however, use Bag of Words encoding instead of VLAD to take advantage of sparse representations for fast retrieval in large-scale databases.

% AMAIA: The figure is explained in section 3. I define it here so that it appears earlier in the paper.
\begin{figure*}[ht]
  \includegraphics[width=\textwidth]{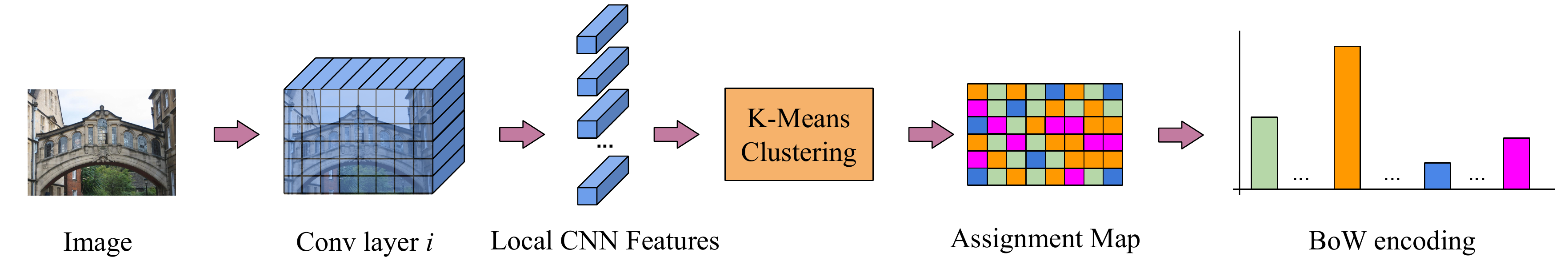}
  \caption{The Bag of Local Convolutional Features pipeline (BLCF).}
  \label{fig:bow}
  \vspace{-10.5pt}
\end{figure*}

% Spatial search with CNNs
Some of the approaches cited above propose systems that are based or partially based on spatial search over multiple regions of the image. Razavian et al.~\cite{razavian2015} achieve a remarkable increase in performance by applying a spatial search strategy over an arbitrary grid of windows at different scales. Although they report high performance in several retrieval benchmarks, their proposed approach is very computationally costly and does not scale well to larger datasets and real-time search scenarios. Tolias et al.~\cite{tolias2015} introduce a local analysis of multiple image patches, which is only applied to the top elements of an initial ranking. They propose an efficient workaround for sub patch feature pooling based on integral images, which allows them to quickly evaluate many image windows. Their approach improves their baseline ranking and also provides approximate object localizations. In this direction, our work proposes using the assignment map to quickly build the BoW representation of any image patch, which allows us to apply a spatial search for reranking. Unlike~\cite{tolias2015}, we use the object localizations obtained with spatial search to mask out the activations of the background and perform query expansion using the detected object location.

\section{Bag of Words Framework}
\label{3_bow}

% CNN architecture / Conv activations as local features

The proposed pipeline for feature extraction uses the activations at different locations of a convolutional layer in a pre-trained CNN as local features. Similar to~\cite{tolias2015,kalantidis2015}, we discard the softmax and fully connected layers of the original network while keeping the original image aspect ratio. Each convolutional layer in the network has $D$ different $N \times M$ feature maps, which can be viewed as $N \times M$ local descriptors of dimension $D$.

% The advantages of BoW
We propose to use bag of words encode the local convolutional features of an image into a single vector. Although more elaborate aggregation strategies have been shown to outperform BoW-based approaches for some tasks in the literature~\cite{vlad,fishervectors}, Bag of words encodings produce sparse high-dimensional codes that can be stored in inverted indices, which are beneficial for fast retrieval. Moreover, BoW-based representations are more compact, faster to compute and easier to interpret.

Bag of words models require constructing a visual codebook to map vectors to their nearest centroid. We use $k$-means on local CNN features to fit this codebook. Each local CNN feature in the convolutional layer is then assigned its closest visual word in the learned codebook. This procedure generates the \textit{assignment map}, i.e. a 2D array of size $N \times M$ that relates each local CNN feature with a visual word. The assignment map is therefore a compact representation of the image which relates each pixel of the original image with its visual word with a precision of $ \left(\frac{W}{N}, \frac{H}{M}\right)$ pixels, where $W$ and $H$ are the width and height of the original image. This property allows us to quickly generate the BoW vectors of not only the full image, but also its parts. We describe the use of this property in our work in Section~\ref{5_retrieval}.

Figure~\ref{fig:bow} shows the pipeline of the proposed approach, which encodes the image into a sparse high dimensional descriptor to be used for instance retrieval.

\section{Image Retrieval}
\label{5_retrieval}

This section describes the image retrieval pipeline, which consists of an initial ranking stage, followed by a spatial reranking, and query expansion.

\textbf{(a) Initial search: }  The initial ranking is computed using the cosine similarity between the BoW vector of the query image and the BoW vectors of the full images in the database. We use a sparse matrix based inverted index and GPU-based sparse matrix multiplications to allow fast retrieval. The image list is then sorted based on the cosine similarity of its elements to the query. We use two types of image search based on the query information that is used:

% \paragraph*{Initial search} The initial ranking is computed using the cosine similarity between the BoW vector of the query image and the BoW vectors of the full images in the database. We use a sparse matrix based inverted index and GPU-based sparse matrix multiplications to allow fast retrieval. The image list is then sorted based on the cosine similarity of its elements to the query. We use two types of image search based on the query information that is used:
\vspace{-2.25pt}

\begin{itemize}

\setlength{\itemsep}{0pt}
\item Global search (GS): The BoW vector of the query is built with the visual words of all the local CNN features in the convolutional layer extracted for the query image.

\item Local search (LS): The BoW vector of the query contains only the visual words of the local CNN features that fall inside the query bounding box.
\end{itemize}
\vspace{-2.25pt}

% \begin{figure}
% \centering
% \includegraphics[width=\columnwidth]
% {figures/LQS.png}
% \caption{Selection of Local CNN features to construct the BoW query descriptor for the Local Search (LS).}
% \label{fig_lqs}
% \end{figure}

\textbf{(b) Local reranking (R): } After the initial search, the top $T$ images in the ranking are reranked based on a localization score. We choose windows of all possible combinations of width $w \in \{W, \frac{W}{2}, \frac{W}{4}\}$ and height $h \in \{H, \frac{H}{2}, \frac{H}{4}\}$, where $W$ and $H$ are the width and height of the assignment map. We use a sliding window strategy directly on the assignment map with 50\% of overlap in both directions. 

%\paragraph*{Local reranking (R)} After the initial search, the top $T$ images in the ranking are locally analyzed and reranked based on a localization score. We choose windows of all possible combinations of width $w \in \{W, \frac{W}{2}, \frac{W}{4}\}$ and height $h \in \{H, \frac{H}{2}, \frac{H}{4}\}$, where $W$ and $H$ are the width and height of the assignment map. We employ a sliding window strategy directly on the assignment map with 50\% of overlap in both directions. 

We additionally perform a simple filtering strategy to discard those windows whose aspect ratio is too different to the aspect ratio of the query. Let the aspect ratio of the query bounding box be $AR_q = \frac{W_q}{H_q}$ and $AR_w = \frac{W_w}{H_w}$ be the aspect ratio of the window. The score for window $w$ is defined as
$score_{w} = \frac{min(AR_w,AR_q)}{max(AR_w,AR_q)}$.
%
% \begin{equation}
% score_{w} = \frac{min(AR_w,AR_q)}{max(AR_w,AR_q)} 
% \end{equation}
%
All windows with a score lower than a threshold $th$ are discarded. 
%
% KEVIN: @Eva, @Amaia, please include the value of this threshold in the paper (reproducability). Also, it would be good to include how you arrived at this value (based on some tests on a validation set?, by visual inspection of some results?). The notation $th_{w}$ makes it look like the value of the threshold is a function of the window. But it is a constant, right? If so, it would be better just to name it something like $t$, or leave it unnamed.

% AMAIA: I replaced $th_{w}$ with simply th. The value of the threshold is provided later in the paper.
%

For each of the remaining windows, we construct the BoW vector representation and compare it with the query representation using cosine similarity. The window with the highest cosine similarity is picked as the object localization and its score is kept for the image. We also enhance the BoW window representation with spatial pyramid matching~\cite{spm} with $L = 2$ resolution levels (i.e. the full window and its 4 sub regions). We construct the BoW representation of all sub regions at the 2 levels, and weight their contribution to the similarity score with inverse proportion to the resolution level of the region. The cosine similarity of a sub region $r$ to the corresponding query sub region is therefore weighted by $ w_{r} = \frac{1}{2^{(L-l_{r})}} $, where $l_{r}$ is the resolution level of the region $r$. With this procedure, the top $T$ elements of the ranking are sorted based on the cosine similarity of their regions to the query's, and also provides the region with the highest score as a rough localization of the object.

% \begin{figure}[H]
% \centering
% \includegraphics[width=\columnwidth]
% {figures/spm.pdf}
% \caption{Spatial Pyramid matching on window locations.}
% \label{fig_spm}
% \end{figure}

\textbf{(c) Query expansion} We investigate two query expansion strategies based on global and local BoW descriptors: 

%\paragraph*{Query expansion} We investigate two query expansion strategies based on global and local BoW descriptors: 

\begin{itemize}
\setlength{\itemsep}{0pt}
\item Global query expansion (GQE): The BoW vectors of the $N$ images at the top of the ranking are averaged together with the BoW of the query to form the new representation for the query.

\item Local query expansion (LQE): Locations obtained in the local reranking step are used to mask out the background and build the BoW descriptor of only the region of interest of the $N$-top images in the ranking. These BoW vectors are averaged together with the BoW of the query bounding box to perform a second search.

\end{itemize}

\section{Experiments}
\label{5_experiments}

\subsection{Preliminary experiments}
We experiment with three different instance retrieval benchmarks to evaluate the performance of our approach: Oxford Buildings \cite{philbin2007object}, Paris Buildings \cite{paris6k} and a subset of TRECVid Instance Search 2013 (INS 23k) \cite{trecvid} containing only those keyframes that are relevant to at least one of the queries. For the Paris and Oxford datasets, we also test the performance when adding 100,000 distractor images collected from Flickr to the original datasets (resulting in the Oxford 105k and Paris 106k datasets, respectively). 

Feature extraction was performed using \emph{Caffe}~\cite{caffe} and the VGG16 pre-trained network~\cite{vgg}. We extracted features from the last three convolutional layers (conv5\_1, conv5\_2 and conv5\_3) and compared their performance on the Oxford 5k dataset. We experimented on different image input sizes: 1/3 and 2/3 of the original image. Following several other authors~\cite{babenko2015,kalantidis2015}, we $l_2$-normalize all local features, followed by PCA, whitening, and a second round of $l_2$-normalization. The PCA models were fit on the same dataset as the test data in all cases. Unless stated otherwise, all experiments used a visual codebook of 25,000 centroids fit using the ($L^2$-PCA-$L^2$ transformed) local CNN features of all images in the same dataset.

We apply bilinear interpolation on the convolutional layers to obtain higher resolution maps as a workaround of using bigger input size image. We find this strategy to be beneficial for all layers tested, achieving the best results when using conv5\_1. Inspired by the boost in performance of the Gaussian center prior in SPoC features~\cite{babenko2015}, we apply a weighting scheme on the visual words to provide more importance to those belonging to the center of the image. 

\subsection{Reranking and query expansion}

We apply the local reranking (R) stage on the top-100 images in the initial ranking, using the sliding window approach described in Section~\ref{5_retrieval}. 
The presented aspect ratio filtering is applied with a threshold $th = 0.4$, which was chosen based on a visual inspection of results on a subset of Oxford 5k. 
%Using this procedure, we locally analyze and rerank the top-100 images from the initial ranking. 
Query expansion is later applied considering the top-10 images of the resulting ranking. 
Table~\ref{q_exp} contains the results for the different stages in the pipeline.

\begin{table}
\centering
\small
\caption{mAP on Oxford 5k and Paris 6k for the different stages in the pipeline introduced in Section~\ref{5_retrieval}.}
\label{q_exp}
\begin{tabular}{@{}ccccccc@{}}
\toprule
                           &  &  Baseline    & +R    & +GQE  & \multicolumn{1}{l}{\begin{tabular}[c]{@{}l@{}}+R\\ +GQE\end{tabular}} & \multicolumn{1}{l}{\begin{tabular}[c]{@{}l@{}}+R\\ +LQE\end{tabular}} \\ \midrule
\multirow{2}{*}{Oxford} & GS    & 0.653 & 0.701 & 0.702 & 0.771                                                                 & \textbf{0.782}                                                        \\
                           & LS    & 0.738 & 0.734 & 0.773 & 0.769                                                                 & \textbf{0.788}                                                        \\
\multirow{2}{*}{Paris}  & GS    & 0.699 & 0.719 & 0.774 & 0.801                                                                 & \textbf{0.835}                                                        \\
                           & LS    & 0.820 & 0.815 & 0.814 & 0.807                                                                 & \textbf{0.848}                                                        \\ \midrule
\end{tabular}
\vspace{-10.5pt}
\end{table}

The results indicate that the local reranking is significantly beneficial only when applied to a ranking obtained from a search using the global BoW descriptor of the query image (GS). This is consistent with the work by Tolias et al.~\cite{tolias2015}, who also apply a spatial reranking followed by query expansion to a ranking obtained with a search using descriptors of full images. They achieve a mAP of 0.66 in Oxford 5k, which is increased to 0.77 after spatial reranking and query expansion, while we reach similar results (e.g. from 0.652 to 0.769). However, our results indicate that a ranking originating from a local search (LS) does not benefit from local reranking. Since the BoW representation allows us to effectively perform a local search (LS) in a database of full indexed images, we find the local reranking stage applied to LS to be redundant in terms of the achieved quality of the ranking. However, the local reranking stage does provide with a rough localization of the object in the images of the ranking, as depicted in Figure~\ref{fig:rankings}. We use this information to perform query expansion based on local features (LQE). 

Results indicate that query expansion stages greatly improve performance in Oxford 5k. We do not observe significant gains after reranking and QE in the Paris 6k dataset, although we achieve our best result with LS + R + LQE.

%In the case of augmented queries (+$Q_{aug}$), we find query expansion to be less helpful in all cases, which suggests that the information gained with query augmentation and the one obtained by means of query expansion strategies are not complementary. 

\subsection{Comparison with the state-of-the-art}

We compare our approach with other CNN-based representations that make use of features from convolutional layers on the Oxford and Paris datasets. Table~\ref{tab_soa} includes the best result for each approach in the literature. Our performance using global search is comparable to that of Ng et al.~\cite{ng2015}, which is the one that most resembles our approach. However, they achieve this result using VLAD features, which are more expensive to compute and, being a dense high-dimensional representation, do not scale as well to larger datasets. Similarly, Razavian et al.~\cite{razavian2015} achieve the highest performance of all approaches in both the Oxford and Paris benchmarks by applying a spatial search at different scales for all images in the database. Such approach is prohibitively costly when dealing with larger datasets, especially for real-time search scenarios. Our BoW-based representation is highly sparse, allowing for fast retrieval using inverted indices, and achieves consistently high mAP in all tested datasets. 

%AMAIA: What to say about our drop in performance in Oxford 105 and Paris 106? Our results are much worse than the rest.
%
% KEVIN: In response to the above -- we could just be honest about it. Perhaps state somewhere in the conclusion that: "Our method does, however, appear to be more sensitive to large numbers of distractor images than methods based on sum and max pooling (SPoC, R-MAC, and CroW). We speculate that this may be because the distractor images are drawn from a different distribution to the original dataset, and may therefore require a larger codebook to better represent the diversity in the visual words. Future work will investigate this further." EDIT: Actually, I updated the conclusion to reflect this.

%We find the usage of the query bounding box to be extremely beneficial in our case for both datasets. The authors of SPoC~\cite{babenko2015} are the only ones who report results using the query bounding box for search, finding a decrease in performance from 0.589 to 0.531 using raw SPoC features (without center prior). This suggests that sum pooled CNN features are less suitable for instance level search in datasets where images are represented with global descriptors. 
% EVA: Not suitable for instance level search without doing spatial search, no? Maybe the thing here is that BoW allows 'mixing' scales. Then, we can compare local regions with global ones. When you mix local and global scale in sumpool it is not really working, but it does work doing spatial search - r-mac guys -.

We also compare our local reranking and query expansion results with similar approaches in the state-of-the-art. The authors of R-MAC~\cite{tolias2015} apply a spatial search for reranking, followed by a query expansion stage, while the authors of CroW~\cite{kalantidis2015} only apply query expansion after the initial search. Our proposed approach also achieves competitive results in this section, achieving the best result for Oxford 5k.

\begin{table}[t]
\centering
\caption{Comparison to state-of-the-art CNN representations (mAP). Results in the lower section consider reranking and/or query expansion.}
\label{tab_soa}
\begin{tabular}{lcccc}
\hline
                      & \multicolumn{2}{c}{Oxford} & \multicolumn{2}{c}{Paris} \\ 
                      & 5k           & 105k        & 6k          & 106k        \\ \hline
Ng \emph{et al.} \cite{ng2015}                    & 0.649        & -           & 0.694       & -           \\ 
Razavian \emph{et al.} \cite{razavian2015}              &\textbf{0.844}& -           & \textbf{0.853}       & -           \\ 
SPoC \cite{babenko2015}                & 0.657        & 0.642       & -           & -           \\ 
R-MAC \cite{tolias2015}                & 0.668        & 0.616       & 0.830       & \textbf{0.757}       \\ 
CroW \cite{kalantidis2015}                 & 0.682        & \textbf{0.632}       & 0.796       & 0.710       \\ 
uCroW \cite{kalantidis2015}                 & 0.666        & 0.629       & 0.767      & 0.695       \\ 
GS            & 0.652        & 0.510      & 0.698       & 0.421       \\ 
LS            & 0.739        & 0.593      & 0.820       & 0.648       \\ 
%LS + $Q_{aug}$       & 0.758        & 0.622     & 0.832       & 0.673      \\
\hline
CroW + GQE \cite{kalantidis2015}            & 0.722        & 0.678       & 0.855       & 0.797       \\ 

R-MAC + R + GQE \cite{tolias2015}      & 0.770        & \textbf{0.726}       & \textbf{0.877}       & \textbf{0.817}       \\ 
LS + GQE      &0.773& 0.602           & 0.814       & 0.632           \\ 
LS + R + LQE & \textbf{0.788} & 0.651           & 0.848       & 0.641           \\ 
%LS + R + GQE + $Q_{aug}$ & \textbf{0.793}& 0.666           & 0.828       & 0.683  \\
\hline
\end{tabular}
\vspace{-10.5pt}
\end{table}

\subsection{Experiments on TRECVid INS}

In this section, we compare the Bag of Local Convolutional Features (BLCF) with the sum pooled convolutional features proposed in several works in the literature. We use our own implementation of the \emph{uCroW} descriptor from \cite{kalantidis2015} and compare it with BLCF for the TRECVid INS subset. For the sake of comparison, we test our implementation of sum pooling using both our chosen CNN layer and input size (conv5\_1 and 1/3 image size), and the ones reported in \cite{kalantidis2015} (pool5 and full image resolution). For the BoW representation, we train the codebook using 3M local CNN features chosen randomly from the INS subset. In this case, we do not apply center prior to the feature to avoid down weighting local features from image areas where the objects might appear. Table \ref{trecvid_table} compares sum pooling with BoW in Oxford, Paris, and TRECVid subset datasets. As stated in earlier sections, sum pooling and BoW have similar performance in Oxford and Paris datasets. For the TRECVid INS subset, however, Bag of Words significantly outperforms sum pooling, which demonstrates its suitability for challenging instance search datasets, in which queries are not centered and have variable size and appearance.  We also observe a different behavior when using the provided query object locations (LS) to search, which was highly beneficial in Oxford and Paris datasets, but does not provide any gain in TRECVid INS. We hypothesize that the fact that the size of the instances is much smaller in TRECVid than in Paris and Oxford datasets causes this drop in performance. Global search (GS) achieves better results on TRECVid INS, which suggests that query instances are in many cases correctly retrieved due to their context.

\begin{table}[ht]
\centering
\caption{mAP of sum pooling and BoW aggregation techniques in Oxford, Paris and TRECVid INS subset.}
\label{trecvid_table}
% \begin{tabular}{@{}ccccccc@{}}
% \toprule
% \multicolumn{2}{l}{\textbf{}}     & \multicolumn{2}{c}{Oxford} & \multicolumn{2}{c}{Paris} & \begin{tabular}[c]{@{}c@{}}INS \\ (subset)\end{tabular} \\ 
% \midrule
% \multicolumn{2}{l}{}              & 5k           & 105k        & 6k          & 106k        &     23k  \\ 
% \midrule
% \multirow{2}{*}{Ours}             & GS & 0.650        & 0.510       & 0.698       & 0.421       & 0.323  \\ 
% 							      & WS & 0.706        & ?           & 0.762       & 0.532       & \textbf{0.353}  \\
%                                   & LS & 0.739        & 0.593       & 0.819       & 0.648       & 0.295  \\ 
% \midrule
% \multirow{2}{*}{Sum pool}         & GS & 0.621        & 0.578       & 0.398       & 0.230       & 0.156  \\  
%                                   & WS & 0.566        & 0.475       & 0.462       & 0.289       & 0.151  \\
%          (as ours)                & LS & 0.572        & 0.349       & 0.410       & 0.230       & 0.097  \\

% \midrule
% \multirow{2}{*}{Sum pool}         & GS & 0.672     &   0.629*   &   0.774      &    0.695*    &  0.139 \\  
%                                   & WS & 0.713     &   --       &   0.796      &    --        &  0.148 \\
% (as in \cite{kalantidis2015})     & LS & 0.683     &    --      &   0.763      &    --        &  0.120 \\

% \bottomrule
% \end{tabular}
\begin{tabular}{@{}ccccc@{}}
\toprule
                                                                                 &  & Oxford 5k & Paris 6k & INS 23k \\ \midrule
\multirow{2}{*}{BoW}                                                             & GS & 0.650     & 0.698    & \textbf{0.323}   \\
                                                                                 %& WS & 0.693     & 0.742    & \textbf{0.350}   \\
                                                                                 & LS & \textbf{0.739}     & \textbf{0.819}    & 0.295   \\ \midrule
\multirow{2}{*}{\begin{tabular}[c]{@{}c@{}}Sum pooling\\ (as ours)\end{tabular}} & GS & 0.606     & 0.712    & 0.156   \\
                                                                                 %& WS & 0.638     & 0.745    & 0.150   \\
                                                                                 & LS & 0.583     & 0.742    & 0.097   \\ \midrule
\multirow{2}{*}{\begin{tabular}[c]{@{}c@{}}Sum pooling\\ (as in \cite{kalantidis2015})\end{tabular}}  & GS & 0.672     & 0.774    & 0.139   \\
                                                                                % & WS & 0.707     & 0.789    & 0.146   \\
                                                                                 & LS & 0.683     & 0.763    & 0.120   \\  \midrule
\end{tabular}
\end{table}

\section{Conclusion}
\label{6_conclusions}

We proposed an aggregation strategy based on Bag of Words to encode features from convolutional neural networks into a sparse representations for instance search. We achieved competitive performance with respect to other CNN-based representations in Oxford and Paris benchmarks, while being more scalable in terms of index size, cost of indexing, and search time. We also compared our BoW encoding scheme with sum pooling of CNN features in the far more challenging TRECVid instance search task, and demonstrated that our method consistently and significantly performs better. Our method does, however, appear to be more sensitive to large numbers of distractor images than methods based on sum and max pooling. We speculate that this may be because the distractor images are drawn from a different distribution to the original dataset, and may therefore require a larger codebook to better represent the diversity in the visual words. 

%We compared the BoW aggregation scheme with sum and max pooling over convolutional feature maps for instance search in complex scenes (TRECVid Instance Search), finding BoW to consistently outperform the other two by a great margin.
\section*{Acknowledgements}
This publication has emanated from research conducted with the financial support of Science Foundation Ireland (SFI) under grant number SFI/12/RC/2289, project BigGraph TEC2013-43935-R, funded by the Spanish Ministerio de Econom\'ia y Competitividad and the European Regional Development Fund (ERDF), as well as the donation of a GeForce GTX Titan X from NVIDIA Corporation.

% KEVIN: We might want to put in an acknowledgements section before camera ready that says we used GPU cards from Nvidia for some of this work

{
\bibliographystyle{ieee}
\bibliography{egbib}

\begin{thebibliography}{10}\itemsep=-1pt

\bibitem{netvlad}
R.~Arandjelovi{\'c}, P.~Gronat, A.~Torii, T.~Pajdla, and J.~Sivic.
\newblock {NetVLAD}: {CNN} architecture for weakly supervised place
  recognition.
\newblock {\em arXiv:1511.07247}, 2015.

\bibitem{babenko2015}
A.~Babenko and V.~Lempitsky.
\newblock Aggregating local deep features for image retrieval.
\newblock In {\em International Conference on Computer Vision (ICCV)}, December
  2015.

\bibitem{neuralcodes}
A.~Babenko, A.~Slesarev, A.~Chigorin, and V.~Lempitsky.
\newblock Neural codes for image retrieval.
\newblock In {\em Computer Vision--ECCV 2014}. 2014.

\bibitem{gong2014multi}
Y.~Gong, L.~Wang, R.~Guo, and S.~Lazebnik.
\newblock Multi-scale orderless pooling of deep convolutional activation
  features.
\newblock In {\em Computer Vision--ECCV 2014}. 2014.

\bibitem{vlad}
H.~J{\'e}gou, M.~Douze, C.~Schmid, and P.~P{\'e}rez.
\newblock Aggregating local descriptors into a compact image representation.
\newblock In {\em Computer Vision and Pattern Recognition (CVPR)}, 2010.

\bibitem{caffe}
Y.~Jia, E.~Shelhamer, J.~Donahue, S.~Karayev, J.~Long, R.~Girshick,
  S.~Guadarrama, and T.~Darrell.
\newblock Caffe: Convolutional architecture for fast feature embedding.
\newblock In {\em ACM Multimedia}, 2014.

\bibitem{kalantidis2015}
Y.~Kalantidis, C.~Mellina, and S.~Osindero.
\newblock Cross-dimensional weighting for aggregated deep convolutional
  features.
\newblock {\em arXiv:1512.04065}, 2015.

\bibitem{spm}
S.~Lazebnik, C.~Schmid, and J.~Ponce.
\newblock Beyond bags of features: Spatial pyramid matching for recognizing
  natural scene categories.
\newblock In {\em Computer Vision and Pattern Recognition (CVPR)}, 2006.

\bibitem{deepindex}
Y.~Liu, Y.~Guo, S.~Wu, and M.~S. Lew.
\newblock Deepindex for accurate and efficient image retrieval.
\newblock In {\em International Conference on Multimedia Retrieval (ICMR)},
  2015.

\bibitem{ng2015}
J.~Ng, F.~Yang, and L.~Davis.
\newblock Exploiting local features from deep networks for image retrieval.
\newblock In {\em Computer Vision and Pattern Recognition Workshops (CVPRW)},
  2015.

\bibitem{nguyen2015query}
V.-T. Nguyen, D.-L. Nguyen, M.-T. Tran, D.-D. Le, D.~A. Duong, and S.~Satoh.
\newblock Query-adaptive late fusion with neural network for instance search.
\newblock In {\em International Workshop on Multimedia Signal Processing
  (MMSP)}, 2015.

\bibitem{fishervectors}
F.~Perronnin, Y.~Liu, J.~S{\'a}nchez, and H.~Poirier.
\newblock Large-scale image retrieval with compressed fisher vectors.
\newblock In {\em Computer Vision and Pattern Recognition (CVPR)}, 2010.

\bibitem{philbin2007object}
J.~Philbin, O.~Chum, M.~Isard, J.~Sivic, and A.~Zisserman.
\newblock Object retrieval with large vocabularies and fast spatial matching.
\newblock In {\em Computer Vision and Pattern Recognition (CVPR)}, 2007.

\bibitem{paris6k}
J.~Philbin, O.~Chum, M.~Isard, J.~Sivic, and A.~Zisserman.
\newblock Lost in quantization: {I}mproving particular object retrieval in
  large scale image databases.
\newblock In {\em Computer Vision and Pattern Recognition (CVPR)}, 2008.

\bibitem{cnnofftheshelf}
A.~S. Razavian, H.~Azizpour, J.~Sullivan, and S.~Carlsson.
\newblock {CNN} features off-the-shelf: an astounding baseline for recognition.
\newblock In {\em Computer Vision and Pattern Recognition Workshops (CVPRW)},
  2014.

\bibitem{razavian2015}
A.~Sharif~Razavian, J.~Sullivan, A.~Maki, and S.~Carlsson.
\newblock A baseline for visual instance retrieval with deep convolutional
  networks.
\newblock In {\em International Conference on Learning Representations}. ICLR,
  2015.

\bibitem{vgg}
K.~Simonyan and A.~Zisserman.
\newblock Very deep convolutional networks for large-scale image recognition.
\newblock {\em arXiv preprint arXiv:1409.1556}, 2014.

\bibitem{bow}
J.~Sivic and A.~Zisserman.
\newblock Efficient visual search of videos cast as text retrieval.
\newblock {\em IEEE Transactions on Pattern Analysis and Machine Intelligence},
  31(4), 2009.

\bibitem{trecvid}
A.~F. Smeaton, P.~Over, and W.~Kraaij.
\newblock Evaluation campaigns and trecvid.
\newblock In {\em International Workshop on Multimedia Information Retrieval
  (MIR)}, 2006.

\bibitem{tolias2015}
G.~Tolias, R.~Sicre, and H.~J{\'e}gou.
\newblock Particular object retrieval with integral max-pooling of {CNN}
  activations.
\newblock {\em arXiv preprint arXiv:1511.05879}, 2015.

\bibitem{xie2015image}
L.~Xie, Q.~Tian, R.~Hong, and B.~Zhang.
\newblock Image classification and retrieval are one.
\newblock In {\em International Conference on Multimedia Retrieval (ICMR)},
  2015.

\bibitem{zhang2015topological}
W.~Zhang and C.-W. Ngo.
\newblock Topological spatial verification for instance search.
\newblock {\em IEEE Transactions on Multimedia}, 17(8), 2015.

\bibitem{zhou2014practical}
X.~Zhou, C.-Z. Zhu, Q.~Zhu, S.~Satoh, and Y.-T. Guo.
\newblock A practical spatial re-ranking method for instance search from
  videos.
\newblock In {\em International Conference on Image Processing (ICIP)}, 2014.

\bibitem{zhu2012large}
C.-Z. Zhu and S.~Satoh.
\newblock Large vocabulary quantization for searching instances from videos.
\newblock In {\em International Conference on Multimedia Retrieval (ICMR)},
  2012.

\end{thebibliography}
}

\end{document}